\documentclass[11pt]{article}

\usepackage[preprint]{acl}
\usepackage{soul}
\usepackage{tikz}

\usepackage{times}
\usepackage{latexsym}
\usepackage{hyperref}
\usepackage{amssymb}
\usepackage{nccmath}
\usepackage{graphicx}

\usepackage[T1]{fontenc}

\usepackage[utf8]{inputenc}

\usepackage{microtype}

\usepackage{inconsolata}

\usepackage{graphicx}
\usepackage{booktabs}
\usepackage{multirow}
\usepackage{array}
\usepackage{amsmath}
\usepackage{listings}
\usepackage{float}
\usepackage{makecell}

\title{Helping Language Models Reason to Access their World Knowledge}
\title{Reinforcement Learning Also Helps Models on Factual Recall Tasks}
\title{Reinforcement Learning Also Helps Models Access Knowledge Better}
\title{Thinking to Remember: Understanding and Improving how Language Models Reason for Knowledge Recall}
\title{Improving Language Model Reasoning Traces for Knowledge Access}
\title{Helping Language Models Reason for Knowledge Recall}
\title{Improving Language Model Reasoning for Knowledge Recall}
\title{Improving Language Model Reasoning for Knowledge Recall}
\title{Improving Parametric Knowledge Access\\in Reasoning Language Models}

\author{Melody Ma \and John Hewitt \\
        Columbia University \\
        \texttt{\{ym3065, jh5020\}@columbia.edu}}

\begin{document}
\maketitle
\begin{abstract}

We study reasoning for accessing world knowledge stored in a language model’s parameters.
For example, recalling that Canberra is Australia’s capital may benefit from thinking through major cities and the concept of purpose-built capitals.
While reasoning language models are trained via reinforcement learning to produce reasoning traces on tasks such as mathematics, they may not reason well for accessing their own world knowledge.
We first find that models do not generate their best world knowledge reasoning by default: adding a simple \textit{think step-by-step} cue demonstrates statistically significant improvement in knowledge recall but not math.
Motivated by this, we propose training models to reason over their parametric knowledge using world-knowledge question answering as a verifiable reward.
After reinforcement learning on TriviaQA (+9.9\%), performance also improves on Natural Questions, HotpotQA, SimpleQA, and StrategyQA by 4.2\%, 2.1\%, 0.6\%, and 3.0\%, respectively.
Reasoning models are under-optimized for parametric knowledge access, but can be easily trained to reason better.

\end{abstract}

\section{Introduction}

Reasoning language models trained with Reinforcement Learning from Verifiable Rewards (RLVR) \cite{cobbe2021training, lightman2023let, openai2024learning, teamqwen2024qwq, guo2025deepseek} have achieved strong performance on mathematics and coding benchmarks.
For example, frontier models achieve near-perfect accuracy on the AIME mathematics competition \cite{openaiIntroducingGPT52} and around 80\% resolution rate on SWE-bench, a benchmark of real-world software engineering tasks \cite{jimenez2023swe}.
\begin{figure}[t]
\centering
\includegraphics[width=\columnwidth]{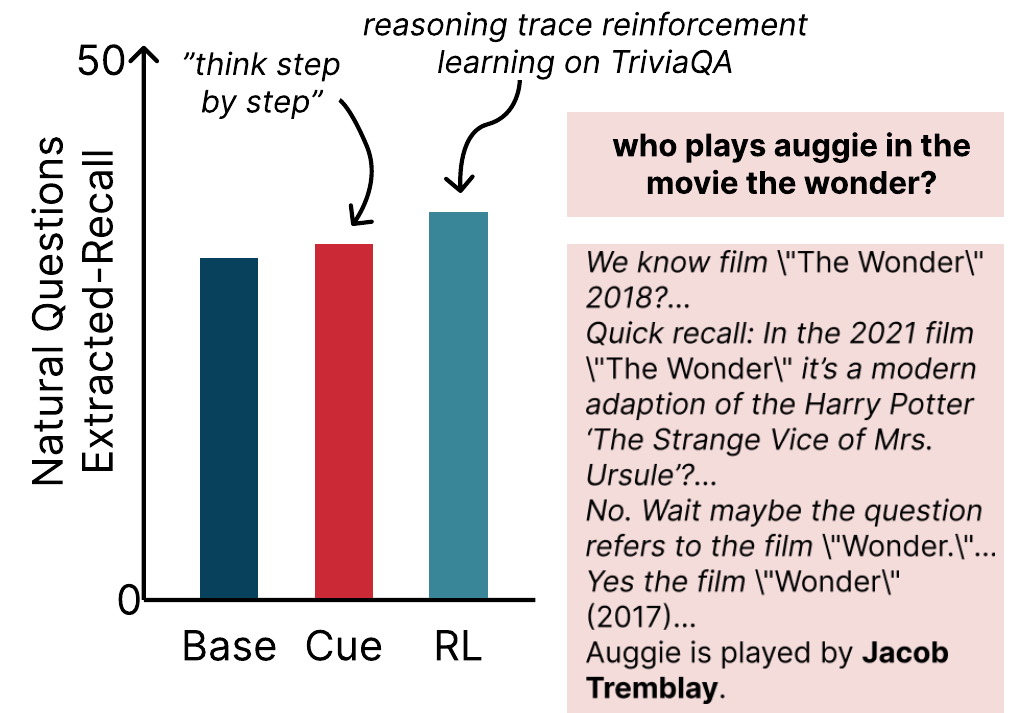}
\caption{\textbf{Left:} GPT-OSS-20B performance: base vs. with \textit{think step-by-step} cue vs. post-RLVR. \textbf{Right:} Sample reasoning trace from the post-RLVR model on a Natural Questions example.}
\label{fig:example}
\end{figure}
Good performance on these tasks requires multi-step reasoning; as a result, models trained on them engage in step-by-step reasoning automatically and often transfer these abilities to other reasoning-heavy domains, such as GPQA \cite{rein2024gpqa} and Humanity's Last Exam \cite{phan2025humanity}.

In this short contribution, we explore the use of reasoning tokens for knowledge recall from model parameters.
We're loosely inspired by spreading activation \cite{collins1975spreading}---where activating one concept in a semantic network causes activation to spread to related concepts.
For example, recalling that Canberra is Australia’s capital may benefit from reasoning through major cities and purpose-built capitals. %
Reasoning to access parametric knowledge is qualitatively different from reasoning used in common RLVR training, and it remains unclear how well reasoning models generate reasoning traces to access their own memory.\footnote{Our code, weights, and evaluation logs are accessible here: \url{https://github.com/MelodyHorsee/parametric-knowledge-access}}

Using two closed-book QA datasets for parametric knowledge retrieval---TriviaQA \cite{joshi-etal-2017-triviaqa} and Natural Questions \cite{kwiatkowski-etal-2019-natural}---we evaluate four reasoning models with and without an explicit \textit{think step-by-step} prompt, which is unnecessary for mathematics or coding. 
Adding the cue modestly but consistently improves knowledge recall---for example, +1.1\% and +1.3\% Ex-Recall \footnote{Defined in Section~\ref{sec:evaluation}} on TriviaQA and Natural Questions for GPT-OSS-20B---but does not help on MATH \cite{hendrycks2021measuring}, where accuracy degrades across three models.
This shows that reasoning language models do not know to do their best knowledge access reasoning by default.

We then propose that RLVR can teach models not only to reason in mathematical, coding, or tool-use domains \cite{wei2025reinforcing, nakano2021webgpt}, but also to better access their \textit{parametric} knowledge.
Training GPT-OSS-20B on TriviaQA with online RL using answer correctness as the reward improves performance on TriviaQA (+27.1\% EM, +9.9\% Ex-Recall), Natural Questions (+12.2\% EM, +4.2\% Ex-Recall), HotpotQA \cite{yang-etal-2018-hotpotqa} (+9.5\% EM, + 2.1\% Ex-Recall), SimpleQA \cite{wei2024measuring} (+1.5\% EM, +0.6\% Ex-Recall), and StrategyQA \cite{geva2021did} (+3.0\% EM).
This also outperforms an offline RL baseline in which we finetune a model on correct reasoning traces generated by the initial model.

After RLVR, we find that reasoning traces are modestly longer (e.g., 94-token average before RLVR for GPT-OSS-20B, 107 after), but it is difficult to qualitatively characterize exactly how they have improved, and anecdotally, the post-RL model seems to sometimes just guess more accurately after brief reasoning.
Intuitively, in facilitating recall from parameters, the ``right'' reasoning is whatever elicits the memory from the model, so reasoning need not be deductive or human-interpretable.

In sum, we show (1) language models do not perform their best reasoning by default on knowledge recall tasks, and (2) RLVR can teach them to better access their parametric knowledge.

\section{Related Work}

\paragraph{RLVR for Reasoning.}
Recent work has demonstrated the effectiveness of RLVR for training language models on mathematics, coding, and other reasoning-intensive tasks, where multi-step reasoning is essential and rewards can be derived from verifiable outcomes \cite{cobbe2021training, lightman2023let, shao2024deepseekmath, guo2025deepseek}.
\paragraph{RL for Open-Book Question Answering.}
RL has been applied to open-book question answering from early retriever-reader systems \cite{wang2018r} to recent work on search-augmented reasoning \cite{jin2025search} and improved RL algorithms \cite{le2025token}. However, all prior work allows access to external knowledge rather than closed-book recall from parametric knowledge.
\paragraph{Closed-Book Knowledge Recall in Language Models.}
Language models store substantial factual knowledge in their parameters \cite{petroni2019language}, enabling closed-book question answering.
Prior work has improved knowledge access through fine-tuning \cite{roberts2020much}, but has not applied reinforcement learning on reasoning traces.

\section{Language Models Can Reason Better on Knowledge Recall Tasks}
\label{sec:language-models-can-reason-better-on-knowledge-recall}

In this section, we demonstrate that the simple \textit{think step-by-step} cue consistently improves model performance on closed-book QA.
We evaluate four reasoning models across two closed-book QA datasets and observe consistent performance improvements --- for example, +1.1\% and +1.3\% Ex-Recall \footnote{Defined in Section~\ref{sec:evaluation}} on TriviaQA and Natural Questions for GPT-OSS-20B.
As a contrast, we find that the same models do not benefit from such prompting on the mathematics benchmark MATH \cite{hendrycks2021measuring}, where they already reason effectively, we hypothesize, thanks to RLVR.
Together, these results indicate that current models do not perform their best reasoning on knowledge recall tasks.

\subsection{Notation and Evaluation Setup}

We denote inputs as $x$ (questions), outputs as $y$ (answers), and reasoning traces as $c$.
Language models are parameterized as $p_\theta$, which first sample a reasoning trace $\hat{c} \sim p_\theta(\cdot | x)$, then generate an answer $\hat{y} \sim p_\theta(\cdot | x, \hat{c})$.
We evaluate two conditions to test whether models automatically perform the best reasoning on knowledge recall tasks:

\indent\textbf{Base:} The input $x$ contains only the question with instructions to provide a final answer.\\
\indent\textbf{+Cue:} The input is augmented with the phrase \textit{``think step-by-step''}, in addition to the base prompt.
Full prompt templates are provided in Appendix~\ref{app:eval_prompts}.

\begin{table}[t]
    \centering
    \small
    \definecolor{CustomBlue}{HTML}{08415C}
    \definecolor{CustomRed}{HTML}{CC2936}
    
    \setlength{\tabcolsep}{2.5pt} 

    \begin{tabular}{@{}ll ccc@{}}
        \toprule
        \textbf{Cue?} & \textbf{Reasoning?} & \textbf{TriviaQA} & \textbf{NQ} & \textbf{MATH} \\
        \midrule
        \multicolumn{5}{@{}l}{\textit{GPT-OSS-20B}} \\
        \textcolor{CustomRed}{$-$\text{Cue}} & \textcolor{CustomRed}{$-$\text{Reasoning}} & 45.2\% & 24.6\% & 25.9\% \\
        \textcolor{CustomRed}{$-$\text{Cue}} & \textcolor{CustomBlue}{$+$\text{Reasoning}} & 60.1\% & 30.7\% & \textbf{80.9\%} \\
        \textcolor{CustomBlue}{$+$\text{Cue}} & \textcolor{CustomBlue}{$+$\text{Reasoning}} & \textbf{61.2\%}$^*$& \textbf{32.0\%}& 80.4\% \\
        \midrule
        \multicolumn{5}{@{}l}{\textit{Olmo-3-7B-Think}} \\
        \textcolor{CustomRed}{$-$\text{Cue}} & \textcolor{CustomRed}{$-$\text{Reasoning}} & 41.4\% & 23.1\% & 71.6\% \\
        \textcolor{CustomRed}{$-$\text{Cue}} & \textcolor{CustomBlue}{$+$\text{Reasoning}} & 55.1\% & 28.0\% & \textbf{85.8\%} \\
        \textcolor{CustomBlue}{$+$\text{Cue}} & \textcolor{CustomBlue}{$+$\text{Reasoning}} & \textbf{56.1\%}$^*$ & \textbf{28.7\%} & 84.8\%$^*$ \\
        \midrule
        \multicolumn{5}{@{}l}{\textit{R1-Distill-Qwen-1.5B}} \\
        \textcolor{CustomRed}{$-$\text{Cue}} & \textcolor{CustomRed}{$-$\text{Reasoning}} & 11.8\% & 6.8\% & \textbf{65.5\%} \\
        \textcolor{CustomRed}{$-$\text{Cue}} & \textcolor{CustomBlue}{$+$\text{Reasoning}} & 12.7\% & 7.4\% & 64.9\% \\
        \textcolor{CustomBlue}{$+$\text{Cue}} & \textcolor{CustomBlue}{$+$\text{Reasoning}} & \textbf{12.8\%} & \textbf{7.6\%} & 64.3\% \\
        \midrule
        \multicolumn{5}{@{}l}{\textit{GPT-5.2}} \\
        \textcolor{CustomRed}{$-$\text{Cue}} & \textcolor{CustomBlue}{$+$\text{Reasoning}} & 90.8\% & 57.0\% & 90.4\% \\
        \textcolor{CustomBlue}{$+$\text{Cue}} & \textcolor{CustomBlue}{$+$\text{Reasoning}} & \textbf{91.5\%}$^*$ & \textbf{58.1\%}$^*$ & \textbf{91.6\%}$^*$ \\
        \bottomrule
    \end{tabular}
\caption{Performance across datasets and models with and without the \textit{think step-by-step} cue and reasoning tokens. NQ = Natural Questions. \textcolor{CustomBlue}{$+$} denotes presence and \textcolor{CustomRed}{$-$} denotes absence. Ex-Recall for TriviaQA and NQ; accuracy for MATH. $*$ indicates the difference between \textcolor{CustomBlue}{$+$}Cue \textcolor{CustomBlue}{$+$}Reasoning and \textcolor{CustomRed}{$-$}Cue \textcolor{CustomBlue}{$+$}Reasoning is statistically significant at the 95\% level by McNemar's test \citep{McNemar_1947}.}
    \label{tab:performance_custom_colors}
\end{table}

\newcolumntype{C}{>{\small\centering\arraybackslash}c}

\subsection{Datasets}
We use three datasets: two closed-book QA datasets for testing knowledge recall and one mathematical reasoning benchmark.
Sample input-output pairs for all three datasets are shown in Appendix~\ref{app:sample_pairs}.

\textbf{TriviaQA} \cite{joshi-etal-2017-triviaqa} is a collection of trivia questions across topics including history, science, and entertainment.
We use the closed-book setting where models answer without supporting documents.
\textbf{Natural Questions} \cite{kwiatkowski-etal-2019-natural} consists of questions from real Google search queries with answers from Wikipedia.
It is also a closed-book benchmark.
\textbf{MATH} \cite{hendrycks2021measuring} contains competition-level 
mathematical reasoning problems.
We include MATH to verify that reasoning models 
already reason effectively on mathematics tasks without explicit prompting.

\subsection{Models}
We evaluate four reasoning models of various sizes and accessibility levels: 
DeepSeek-R1-Distill-Qwen-1.5B (open-weight, 1.5B parameters), Olmo-3-7B-Think (open-weight, 7B parameters), GPT-OSS-20B (open-weight, 20B parameters), and GPT-5.2 (closed-source).
We follow commonly used decoding settings for each model.
Maximum token limits are set based on typical reasoning trace lengths.
For verbose models, we apply budget forcing \cite{muennighoff2025s1}.
Complete hyperparameter settings are in 
Appendix~\ref{app:decoding}.

\subsection{Evaluation}
\label{sec:evaluation}
For TriviaQA and Natural Questions, we evaluate using Exact Match (EM) and Extracted-Recall (Ex-Recall).  
EM measures whether the model's normalized extracted answer matches any normalized reference answer.
Ex-Recall is a slightly relaxed Exact Match. A separate language model (GPT-5-mini) is used to extract a single answer span from the predicted text. We then measure whether any reference answer appears in the extracted span.
We report this since models tend not to respond with a single answer phrase, but computing a recall-like measurement could be gamed by models generating many guesses; the extraction step largely avoids this. The extraction prompt is provided in Appendix~\ref{app:extraction_prompts}.
For MATH, we evaluate accuracy on the final boxed answer.

\subsection{Results}
We find that prompting models with \textit{think step-by-step} leads the models to generate reasoning traces that improve knowledge recall across datasets and models (Table~\ref{tab:performance_custom_colors}).
Across two datasets and four models, we observe an average increase of 0.8\% in Ex-Recall.
As a sanity check, we also verify that reasoning helps knowledge access compared to direct answering. To do this, we modify the prompt format to force the model to emit an empty \texttt{<think>} span. We find that models perform 6.9\% worse on average than with default reasoning.

These results indicate that current language models do not perform their best reasoning on knowledge recall tasks, suggesting their reasoning abilities in these domains are not saturated from RLVR on mathematics and code.

\begin{table*}[t]  %
\centering
\small
\begin{tabular}{lcccccc}
\toprule
& & \textbf{TriviaQA} & \textbf{NQ} & \textbf{HotpotQA} & \textbf{SimpleQA} & \textbf{StrategyQA} \\
\textbf{Model} & \textbf{Metric} & & & & \\
\midrule
\multirow{2}{*}{Base} 
  & EM & 36.5 & 6.0 & 7.5 & 1.8 & 71.5 \\
  & Ex-Recall & 60.1 & 30.7 & 25.5 & 3.5 & -- \\
\midrule
\multirow{2}{*}{Reasoning-SFT } 
  & EM & 38.8 & 6.1 & 7.3 & 1.9 & 72.5 \\
  & Ex-Recall & 63.8 & 31.5 & 26.0 & 4.0 & -- \\
\midrule
\multirow{2}{*}{RL-trained (ours)} 
  & EM & \bf 63.6$^*$ & \bf 18.2$^*$ & \bf 17.0$^*$ & \bf 3.3$^*$ & \bf 74.5$^*$ \\
  & Ex-Recall & \bf 70.0$^*$ & \bf 34.9$^*$ & \bf 27.6$^*$ & \bf 4.1 & -- \\
\bottomrule
\end{tabular}
\caption{GPT-OSS-20B performance before and after RL training on knowledge recall tasks. For our RL-trained model, $^*$ indicates the improvement over the base model is statistically significant at the 95\% level by McNemar's test.}
\label{tab:rl_results}
\end{table*}

\section{Reinforcement Learning for Knowledge Recall Tasks}
\label{sec:rl-for-knowledge-recall-tasks}

Motivated by the findings in Section~\ref{sec:language-models-can-reason-better-on-knowledge-recall}, we investigate reinforcement learning to improve reasoning for
knowledge recall.
We train GPT-OSS-20B using answer correctness as the verifiable reward on TriviaQA \cite{joshi-etal-2017-triviaqa} and evaluate its performance on the TriviaQA, Natural Questions \cite{kwiatkowski-etal-2019-natural}, HotpotQA \cite{yang-etal-2018-hotpotqa}, and SimpleQA \cite{wei2024measuring} test sets, as well as the StrategyQA \cite{geva2021did} training set (which we use due to the small size of its test set).

\subsection{RLVR Formulation}

Our RLVR setup is as follows.
Given an input $x$, the model first generates a reasoning trace
$\hat{c} \sim~p_\theta(\cdot\mid x)$, followed by a final answer
$\hat{y} \sim p_\theta(\cdot \mid x, \hat{c})$.
We train the model using reinforcement learning with a scalar reward
$r(\hat{y}, y)$ that depends only on the final answer and does not directly
supervise the reasoning trace.
The RLVR objective maximizes the expected reward:
\begin{equation}
\max_{\theta} \;
\mathbb{E}_{\substack{x,\, y \\ \hat{c},\, \hat{y} \sim p_\theta(\cdot \mid x)}}
\left[ r(\hat{y},\, y) \right]
\end{equation}

\paragraph{Reward Function.}
We define the reward as
\begin{equation}
r(\hat{y}, y) = r_{\text{answer}}(\hat{y}, y)
+ 0.1 \cdot \bigl(f_{\text{format}}(\hat{y}) - 1\bigr),
\end{equation}
where
\[
r_{\text{answer}}(\hat{y}, y) =
\begin{cases}
1.0 & \text{if } \hat{y} = y \quad \text{(Exact Match)}, \\
0.5 & \text{if } y \subseteq \hat{y} \quad \text{(Recall)}, \\
0 & \text{otherwise},
\end{cases}
\]
and $f_{\text{format}}(\hat{y}) = 1$ if the output correctly uses
\texttt{<answer></answer>} tags and $0$ otherwise.
This reward encourages correct answers—prioritizing exact matches—while
penalizing invalid output formats.

\paragraph{Optimization.}
We optimize the objective using a GRPO-style \cite{shao2024deepseekmath} importance-sampling
policy gradient method.
For each input $x$, we sample a group of $K$ trajectories
$\{(\hat{c}_k, \hat{y}_k)\}_{k=1}^K$ using the current policy and compute
rewards $r_k = r(\hat{y}_k, y)$.
Advantages are computed relative to the group-average reward,
$A_k = r_k - \frac{1}{K}\sum_j r_j$.
The policy gradient is estimated using importance sampling,
\begin{equation}
\begin{aligned}
\scalebox{0.7}{$ %
    \nabla_\theta \mathcal{J}
    = \mathbb{E}\!\left[
    \sum_{k} A_k
    \sum_{i \in \tau_k}
    \frac{\pi_\theta(t_i \mid t_{<i}, x)}
         {\pi_{\theta_{\text{old}}}(t_i \mid t_{<i}, x)}
    \,
    \nabla_\theta \log \pi_\theta(t_i \mid t_{<i}, x)
    \right]
$}
\end{aligned}
\end{equation}
which in our on-policy setting yields importance weights close to one.

\subsection{Training Setup}

\paragraph{Data.}
We use TriviaQA \cite{joshi-etal-2017-triviaqa} for RL training. Since the original dataset does not provide a standard train/validation split, we randomly split the training set 80/20 to create our training and validation sets.

\paragraph{Hyperparameters.}
We conduct online RL training using Tinker\footnote{\url{https://tinker-docs.thinkingmachines.ai/}} with LoRA \cite{hu2021lora} adaptation (rank=32). We use a group size of 8 with 32 groups per batch, learning rate of $2\times 10^{-5}$, KL penalty coefficient of 0.01, and maximum sequence length of 1,028 tokens.

\subsection{Baselines}
To isolate the improvement from improved reasoning, we run two baselines that ablate elements of the RL-for-reasoning process.
\paragraph{Reasoning-SFT.} First, we ablate the on-policy adaptation of the RLVR process, but keep the language model-generated reasoning chains.
In Reasoning-SFT, we generate reasoning traces from the initial GPT-OSS model $p_{\theta_0}$ for examples in a training set.
We then filter to only those for which the model generates the correct answer, and maximize the likelihood of the reasoning traces and correct answers.
Relative to RLVR, this removes the adaptation of the reasoning generation throughout the training process.
See Appendix~\ref{app:reasoning-sft-baseline} for setup details.
\begin{align}
\min_\theta \mathbb{E}_{\substack{x \\ \hat{c},\hat{y} \sim p_{\theta_{0}}(\cdot\mid x)}}
\left[-\log p_\theta(\hat{y}, \hat{c}\mid x)\right]
\end{align}

\paragraph{SFT.} Next, we also ablate the reasoning tokens themselves, running standard supervised tuning on TriviaQA: we condition on the questions (with a boilerplate reasoning filler) and optimize for the negative-log-likelihood of the correct answer, as in \cite{roberts2020much}. See Appendix~\ref{app:sftbaseline} for setup details and results.
\begin{align}
\min_\theta \mathbb{E}_{x,y}\left[-\log p_\theta(y\mid x)\right]
\end{align}

\subsection{Evaluation \& Results}

\paragraph{Datasets.} We evaluate the trained model (checkpoint step 1240) using the same no-cue prompt from Section~\ref{sec:language-models-can-reason-better-on-knowledge-recall} on the test sets of TriviaQA \cite{joshi-etal-2017-triviaqa}, Natural Questions \cite{kwiatkowski-etal-2019-natural}, HotpotQA \cite{yang-etal-2018-hotpotqa}, and SimpleQA \cite{wei2024measuring}. HotpotQA is a multi-hop QA dataset; SimpleQA is a challenging factual QA benchmark designed to evaluate factual accuracy. We use the StrategyQA training set, as the development set is too small to yield reliable estimates.

\paragraph{Results.} We first evaluate the model trained on Reasoning-SFT, which shows consistent improvement over the base model. However, our RL-trained model demonstrates substantially larger gains, as shown in Table~\ref{tab:rl_results}, indicating that RL training contributes beyond what supervised fine-tuning on correct reasoning traces alone provides. 
For example, RL training increases EM on HotpotQA from 7.5\% to 17.0\%, and Ex-Recall from 25.5\% to 27.6\%, 
showing that language models can reason more effectively on knowledge recall tasks. Notably, even after RL training, prompting with the \textit{think step-by-step} cue yields additional gains (Appendix~\ref{app:postrl}), indicating room for further improvement in reasoning for parametric knowledge access.

We also evaluate the trained model on MATH both with and without the \textit{think step-by-step} cue setting. Interestingly, the trained model shows improvement over the base model in the no-cue setting (Appendix~\ref{app:postrl-math}). This transfer of improvement to MATH presents an opportunity for further study.

\paragraph{Analyzing reasoning traces.} Reasoning traces generated by the RL-trained model are consistently longer (Appendix~\ref{app:thinking-length}).
However, the model's internal reasoning does not appear consistently improved when examined qualitatively.
In many cases where the trained model answers correctly and the base model fails, the trained model presents the correct answer earlier in the reasoning trace rather than performing deeper step-by-step reasoning (see Appendix~\ref{app:example_reasoning_traces} for examples).
We view improved answer calibration and improved reasoning as lying on a spectrum, rather than being cleanly separable phenomena: even arriving at the answer earlier may reflect the model correctly recognizing that extended reasoning is unnecessary.
Nonetheless, eliciting qualitatively richer reasoning---such as spreading activation-style knowledge retrieval---remains an open problem.

\section{Conclusion}
This short contribution demonstrates that, first, prompting models with the simple \textit{think step-by-step} cue improves performance on knowledge recall tasks, but not on mathematics, which models have been trained on via reinforcement learning.
Second, we train models on TriviaQA using reinforcement learning with answer correctness as a verifiable reward, observing improved performance across multiple closed-book QA test sets.
In sum, current language models do not perform their best accessing their parametric knowledge, but they can be effectively trained to reason better.

\section*{Limitations}

As discussed in Section~\ref{sec:rl-for-knowledge-recall-tasks}, the reasoning traces produced by the trained models do not reflect improved reasoning in a human-interpretable sense.
An important direction for future work is to develop RLVR objectives that explicitly encourage better reasoning traces, for example by leveraging spreading-activation–style reasoning. Such improvements could potentially lead to further gains in EM and Ex-Recall, particularly on SimpleQA~\cite{wei2024measuring}.

\section*{Acknowledgements}
The authors would like to thank Thinking Machines Lab for a generous grant of Tinker credits, as well as Nick Deas and Lorena Yan for their helpful feedback on drafts of this paper.

\bibliography{custom}

\raggedbottom
\appendix

\section{Prompt Templates}
\label{app:eval_prompts}

We provide the full prompt templates used in our experiments below.

\subsection{TriviaQA and Natural Questions}

\paragraph{Without Cue}

\begin{verbatim}
You will be given a question.
Give your final answer in
<answer></answer> tags.
\end{verbatim}

\paragraph{With Cue}

\begin{verbatim}
You will be given a question.
Think step-by-step and give your final
answer in <answer></answer> tags.
\end{verbatim}

\subsection{MATH}

\paragraph{Without Cue}

\begin{verbatim}
You will be given a question.
Give your final answer in \\boxed{}.
\end{verbatim}

\paragraph{With Cue}

\begin{verbatim}
You will be given a question.
Think step-by-step and give your final answer
in \\boxed{}.
\end{verbatim}

\section{Sample Input-Output Pairs}
\label{app:sample_pairs}

\begin{table}[H]
\centering
\small
\begin{tabular}{lp{0.6\columnwidth}}
\toprule
\textbf{Dataset} & \textbf{Example} \\
\midrule
TriviaQA & \textbf{Q:} In which country was the first permanent bungee jumping site situated? \\
& \textbf{A:} New Zealand \\
\midrule
Natural Questions & \textbf{Q:} When was Harry Potter and the Philosopher's Stone published? \\
& \textbf{A:} 1997 \\
\midrule
MATH & \textbf{Q:} If $(2x+5)(x-3)=14$, find the sum of the possible values of $x$. \\
& \textbf{A:} $\frac{1}{2}$ \\
\bottomrule
\end{tabular}
\caption{Example questions and answers from each dataset.}
\label{tab:dataset_examples}
\end{table}

\section{Decoding Hyperparameters}
\label{app:decoding}

\begin{table}[h]
\centering
\small
\begin{tabular}{lcc}
\toprule
Model & Temperature & Top-$p$ \\
\midrule
DeepSeek-R1-Distill-Qwen-1.5B & 0.6 & 0.95 \\
Olmo-3-7B-Think               & 0.6 & 0.95 \\
GPT-OSS-20B                  & 1.0 & 1.0  \\
\bottomrule
\end{tabular}

\vspace{0.5em}
\begin{minipage}{\linewidth}
\footnotesize
$^\ast$ For GPT-5.2, temperature and top-$p$ could not be explicitly set because we specify
\texttt{reasoning effort = medium}.
\end{minipage}

\caption{Decoding hyperparameters used for each model.}
\label{tab:sampling_settings}
\end{table}

\lstset{
    basicstyle=\ttfamily\small,
    breaklines=true,
    frame=single,
    xleftmargin=0.2cm,
    xrightmargin=0.2cm
}

\section{Auxiliary LLM Extraction Prompt}
\label{app:extraction_prompts}

\begin{lstlisting}[basicstyle=\ttfamily\small, breakindent=0pt, breakautoindent=false]
You are given an answer that may contain one or multiple possibilities.
If it only contains one, just output it as is. Otherwise, choose the answer that is stated with the most confidence, if there are multiple options.
DO NOT correct the answer, even if you think it's incorrect.

Examples:
A: While Leif Erikson reached North America earlier, Christopher Columbus is usually cited.
Refined Answer: Christopher Columbus
A: While some might think Saturn, the largest planet is Jupiter.
Refined Answer: Jupiter
A: It could be Paris, but some might mistakenly say Lyon.
Refined Answer: Paris
A: Leonardo da Vinci painted the Mona Lisa.
Refined Answer: Leonardo da Vinci painted the Mona Lisa.
A: Shanghai is the capital of China.
Refined Answer: Shanghai is the capital of China.
Original Answer: {answer}
Refined Answer: 
\end{lstlisting}

\section{Example Reasoning Trace}
\label{app:example_reasoning_traces}

\noindent\textbf{Question:} Who was Canada's first woman Prime Minister?

\vspace{0.5em}
\noindent\textbf{Base (GPT-OSS-20B, not recalled):}
\begin{quote}
\small\ttfamily
<|channel|>analysis<|message|>This is false: Canada has no woman PM. So answer: none.<|end|><|start|>assistant<|channel|>
final<|message|><answer>Canada has never had a woman Prime Minister.</answer>
\end{quote}

\noindent\textbf{RL-trained (EM):}
\begin{quote}
\small\ttfamily
<|channel|>analysis<|message|>The answer: Kim Campbell? Actually first female PM of Canada: Kim Campbell served in 1993. But also there was female PM? The question: "Canada's first woman Prime Minister" refers to Kim Campbell. Provide answer.<|end|><|start|>assistant<|channel|>
final<|message|><answer>Kim Campbell</answer>
\end{quote}

\noindent\textbf{Question:} The Manuel M.Torres Dam is the largest hydro plant in which country?

\vspace{0.5em}
\noindent\textbf{Base (GPT-OSS-20B, not recalled):}
\begin{quote}
\small\ttfamily
<|channel|>analysis<|message|>Look up: Manuel M. Torres Dam is in South Africa, near the Vaal River. So largest hydro plant in Africa? Wait it's not largest in Africa. Actually the question: largest hydro plant in which country? It is in South Africa. So answer: South Africa.<|end|><|start|>assistant<|channel|>
final<|message|><answer>South Africa</answer>
\end{quote}

\noindent\textbf{RL-trained (EM):}
\begin{quote}
\small\ttfamily
<|channel|>analysis<|message|>We need answer: Manuel M. Torres Dam. It's in Mexico? Wait, Manuel M. Torres is dam in Yucatan? Actually the largest hydroelectric plant in Mexico is the Manuel Moreno Torres Dam (Hidroelectrico Manuel Moreno Torres) on the Grijalva River? Wait that dam? Let's recall: Franceschi? The largest hydroelectric power plant in Mexico is the Chautla, but the biggest dam is the Cerro Prieto? But the Manuel M. Torres is on the Cauvery? I think it's in Venezuela? Let's quickly recall: The Manuel Moren Tor. Wait maybe it's the \textbackslash"Manuel M. Torres\textbackslash" in Vietnam? but unknown. Let's search memory: Dr. Manuel M. Torres Dam is the largest hydro plant in Colombia? Let's think: In Colombia there is the \textbackslash"Manuel M. Torres Dam\textbackslash" on the Magdalena? Actually maybe it's in Mexico: The \textbackslash"Manuel Moreno Torres\textbackslash" hydroelectric plant (Manuel Moren?) is the largest hydro plant in Mexico (largest by capacity). It's located on the Grijalva River in Chiapas. So answer Mexico.<|end|><|start|>assistant<|channel|>
final<|message|><answer>Mexico</answer>
\end{quote}

\noindent\textbf{Question:} Which writer's first works were gathered into a series of short stories called "Plain Tales From The Hills" in the 1880s?

\vspace{0.5em}
\noindent\textbf{Base (GPT-OSS-20B, not recalled):}
\begin{quote}
\small\ttfamily
<|channel|>analysis<|message|>We recall Plain Tales from the Hills is by Rabindranath Tagore? Wait Tagore's first collection was \textbackslash"Bhuvaneshwar\textbackslash" etc. Plain Tales from the Hills is by \textbackslash"Abdul Hamid\textbackslash"?? Actually it is by \textbackslash"Abdul Hamid\textbackslash"? Wait I think it's by \textbackslash"Abdul Hamid\textbackslash", a Pakistani writer. Let's check: Plain Tales from the Hills (1883) by \textbackslash"Abdul Hamid\textbackslash". Yes.
Let's answer.<|end|><|start|>assistant<|channel|>
final<|message|><answer>Abdul Hamid</answer>
\end{quote}

\noindent\textbf{RL-trained (EM):}
\begin{quote}
\small\ttfamily
<|channel|>analysis<|message|>Answer: Rudyard Kipling.<|end|><|start|>assistant
<|channel|>final<|message|><answer>Rudyard Kipling</answer>
\end{quote}

\section{GPT-OSS-20B Performance After RL Training on Knowledge Access}
\label{app:postrl}

\begin{table}[H]
\centering
\scriptsize
\setlength{\tabcolsep}{1pt}
\begin{tabular}{lcccccc}
\toprule
\textbf{Model} & \textbf{Metric} & \textbf{TriviaQA} & \textbf{NQ} & \textbf{HotpotQA} & \textbf{SimpleQA} & \textbf{StrategyQA} \\
\midrule
\multirow{2}{*}{\makecell[l]{RL-trained\\(no-cues)}} 
  & EM & 63.6 & 18.2 & 17.0 & 3.3 & 74.5 \\
  & Ex-Recall & 70.0 & 34.9 & 27.6 & 4.1 & -- \\
\midrule
\multirow{2}{*}{\makecell[l]{RL-trained\\(with-cues)}} 
  & EM & 62.1 & 15.1 & 14.6 & 3.1 & 74.3 \\
  & Ex-Recall & 70.6 & 36.5 & 28.2 & 4.2 & -- \\
\midrule
\multirow{2}{*}{\textit{Delta}} 
  & \textit{EM} & \textit{-1.5$^*$} & \textit{-3.1$^*$} & \textit{-2.4$^*$} & \textit{-0.2} & \textit{-0.2} \\
  & \textit{Ex-Recall} & \textit{+0.6$^*$} & \textit{+1.6$^*$} & \textit{+0.6} & \textit{+0.1} & -- \\
\bottomrule
\end{tabular}
\caption{GPT-OSS-20B performance after RL training with and without the \textit{think step-by-step} cue on knowledge access datasets. $^*$ indicates the delta is statistically significant at the 95\% level by McNemar's test. }
\label{tab:postrl}
\end{table}

\section{GPT-OSS-20B Performance Before and After RL Training on MATH}
\label{app:postrl-math}

\begin{table}[H]
    \centering
    \small
    \definecolor{CustomBlue}{HTML}{08415C}
    \definecolor{CustomRed}{HTML}{CC2936}
    
    \setlength{\tabcolsep}{2.5pt} 

    \begin{tabular}{@{}ll c@{}}
        \toprule
        \textbf{Cue?} & \textbf{Reasoning?} & \textbf{Accuracy} \\
        \midrule
        \multicolumn{3}{@{}l}{\textit{Base}} \\
        \textcolor{CustomRed}{$-$\text{Cue}} & \textcolor{CustomBlue}{$+$\text{Reasoning}} & 80.9\% \\
        \textcolor{CustomBlue}{$+$\text{Cue}} & \textcolor{CustomBlue}{$+$\text{Reasoning}} & 80.4\% \\
        \midrule
        \multicolumn{3}{@{}l}{\textit{RL-trained}} \\
        \textcolor{CustomRed}{$-$\text{Cue}} & \textcolor{CustomBlue}{$+$\text{Reasoning}} & 83.9\% \\
        \textcolor{CustomBlue}{$+$\text{Cue}} & \textcolor{CustomBlue}{$+$\text{Reasoning}} & 80.4\% \\
        \bottomrule
    \end{tabular}
\caption{GPT-OSS-20B accuracy on MATH before and after RL training. A statistically significant improvement is observed in the no-cue setting (p < 0.05) by McNemar's test.}
\label{tab:postrl}
\end{table}

\section{Reasoning-SFT Baseline Details}
\label{app:reasoning-sft-baseline}
To make our Reasoning-SFT training comparable to RL training, we randomly sample 40,000 (1,240 steps $\times$ 32 batch size) traces with correct recall from the base model's generations. We train with LoRA (rank=32), a learning rate of $1\times 10^{-5}$, and a batch size of 32 for 8 epochs. We select the checkpoint at step 6,000 for evaluation based on validation loss.

\section{SFT Baseline Details}
\label{app:sftbaseline}
For our supervised finetuning of GPT-OSS-20B on TriviaQA, we fix the same TriviaQA dataset as used for reinforcement learning. Likewise, we use the same LoRA hyperparameters (rank 32).
We test three configurations of dataset format, batch size, and learning rate, detailed below.
\begin{lstlisting}[basicstyle=\ttfamily\small, xleftmargin=0pt, framexleftmargin=0pt, breakindent=0pt, breakautoindent=false]
<|channel|>analysis<|message|>Answer:
{answer}.<|end|><|start|>assistant<|channel|>final<|message|><answer>
{answer}</answer>.
\end{lstlisting}
\noindent \textit{learning rate: 1e-4, batch size: 128, epoch: 1} \\
\noindent \textbf{EM:} 21.6\%, \textbf{Recall:} 26.2\%
\\
\begin{lstlisting}[basicstyle=\ttfamily\small, xleftmargin=0pt, framexleftmargin=0pt, breakindent=0pt, breakautoindent=false]
<|channel|>analysis<|message|>Need answer: {answer}.<|end|><|start|>assistant<|channel|>final<|message|>The answer is <answer>{answer}</answer>.
\end{lstlisting}
\noindent \textit{learning rate: 1e-4, Batch size: 128, epoch: 1} \\
\noindent \textbf{EM:} 21.6\%, \textbf{Recall:} 26.3\%
\begin{lstlisting}[basicstyle=\ttfamily\small, xleftmargin=0pt, framexleftmargin=0pt, breakindent=0pt, breakautoindent=false]
<|channel|>analysis<|message|>Need answer: {answer}.<|end|><|start|>assistant<|channel|>final<|message|>The answer is <answer>{answer}</answer>.
\end{lstlisting}
\noindent \textit{learning rate: 2e-5, Batch size: 512, epoch: 5} \\
\noindent \textbf{EM:} 20.7\%, \textbf{Recall:} 25.2\% \\
\noindent \textit{Note: We report Recall instead of Ex-Recall here because the SFT-trained model outputs only a single answer in the \texttt{<answer>...</answer>} tags.}

\section{Thinking Token Length Comparisons}
\label{app:thinking-length}

\begin{table}[H]
\footnotesize
\centering
\setlength{\tabcolsep}{3pt}
\begin{tabular}{llccc}
\toprule
Dataset & Model & Avg. & Recalled & Non-Recalled \\
\midrule
\multirow{4}{*}{TriviaQA} 
 & Base & 94.11 & 41.04 & 174.13 \\
 & SFT & 103.66 & 46.45 & 204.60 \\
 & RL & 106.71 & 55.41 & 226.66 \\
 & RL+cues & 118.65 & 62.02 & 254.88 \\
\midrule
\multirow{4}{*}{NQ} 
 & Base & 90.56 & 44.70 & 110.93 \\
 & SFT & 94.60 & 46.38 & 116.77 \\
 & RL & 104.21 & 58.22 & 128.84 \\
 & RL+cues & 115.64 & 67.32 & 143.42 \\
\midrule
\multirow{4}{*}{HotpotQA} 
 & Base & 160.72 & 80.07 & 188.34 \\
 & SFT & 172.26 & 84.00 & 203.24 \\
 & RL & 188.46 & 103.02 & 221.01 \\
 & RL+cues & 210.84 & 111.58 & 249.77 \\
\midrule
\multirow{4}{*}{SimpleQA} 
 & Base & 124.69 & 77.74 & 126.37 \\
 & SFT & 135.61 & 88.34 & 137.56 \\
 & RL & 148.84 & 116.11 & 150.22 \\
 & RL+cues & 165.06 & 124.15 & 166.87 \\
\midrule
\multirow{3}{*}{StrategyQA} 
 & Base & 43.00 & 40.28 & 49.83 \\
 & SFT & 44.44 & 40.46 & 55.16 \\
 & RL & 52.15 & 46.69 & 67.88 \\
 & RL+cues & 57.91 & 54.21 & 68.62 \\
\bottomrule
\end{tabular}
\caption{Average thinking token length across datasets for GPT-OSS-20B variants. SFT here refers to Reasoning-SFT.}
\label{tab:thinking-length}
\end{table}

\end{document}